\documentclass[letterpaper, 10 pt, conference]{ieeeconf}  

\IEEEoverridecommandlockouts                              

\usepackage{cite}
\usepackage{amsmath,amssymb,amsfonts}
\usepackage{algorithmic}
\usepackage{graphicx}
\usepackage{textcomp}
\usepackage{xcolor}
\usepackage{xspace}
\usepackage{balance}
\usepackage{hyperref}
\usepackage{cleveref}
\usepackage{subcaption}
\usepackage{tcolorbox}
\usepackage{tikz} 
\usepackage{siunitx}
\usepackage{booktabs}
\usepackage{multirow}
\usepackage{etoolbox}
\newcommand{\ubold}{\fontseries{b}\selectfont}
\usepackage{tikz} 
\usepackage{eso-pic} 

\newcommand{\linebreakand}{%
  \end{@IEEEauthorhalign}
  \hfill\mbox{}\par
  \mbox{}\hfill\begin{@IEEEauthorhalign}
}

\robustify\ubold

\title{\LARGE \bf
Robust Reinforcement Learning-Based Locomotion for Resource-Constrained Quadrupeds with Exteroceptive Sensing
}

\author{Davide Plozza$^{1}$, Patricia Apostol$^{1}$, Paul Joseph$^{1}$, Simon Schläpfer$^{1}$, Michele Magno$^{1}$ 
\thanks{$^{1}$Center for Project-Based Learning, ETH Zurich, Zurich, Switzerland}
}

\begin{document}

\maketitle
\thispagestyle{empty}
\pagestyle{empty}

\AddToShipoutPictureBG*{
  \AtPageUpperLeft{%
    \put(0,-40){\raisebox{15pt}{\makebox[\paperwidth]{\begin{minipage}{21cm}\centering
      \textcolor{gray}{ This article has been accepted for publication in the proceedings of the \\
      2025 IEEE International Conference on Robotics and Automation (ICRA) \\
   DOI: 10.1109/ICRA55743.2025.11128474
    } 
      
    \end{minipage}}}}%
  }
  \AtPageLowerLeft{%
    \raisebox{25pt}{\makebox[\paperwidth]{\begin{minipage}{21cm}\centering
      \textcolor{gray}{ \copyright 2025 IEEE.  Personal use of this material is permitted.  Permission from IEEE must be obtained for all other uses, in any current or future media, including reprinting/republishing this material for advertising or promotional purposes, creating new collective works, for resale or redistribution to servers or lists, or reuse of any copyrighted component of this work in other works.
      }
    \end{minipage}}}%
  }
}



\begin{abstract}

Compact quadrupedal robots are proving increasingly suitable for deployment in real-world scenarios. Their smaller size fosters easy integration into human environments.
Nevertheless, real-time locomotion on uneven terrains remains challenging, particularly due to the high computational demands of terrain perception.
This paper presents a robust reinforcement learning-based exteroceptive locomotion controller for resource-constrained small-scale quadrupeds in challenging terrains, which exploits real-time elevation mapping, supported by a careful depth sensor selection.
We concurrently train both a policy and a state estimator, which together provide an odometry source for elevation mapping, optionally fused with visual-inertial odometry (VIO). We demonstrate the importance of positioning an additional time-of-flight sensor for maintaining robustness even without VIO, thus having the potential to free up computational resources. We experimentally demonstrate that the proposed controller can flawlessly traverse steps up to 17.5 cm in height and achieve an 80\% success rate on 22.5 cm steps, both with and without VIO.
The proposed controller also achieves accurate forward and yaw velocity tracking of up to 1.0 m/s and 1.5 rad/s respectively. 
We open-source our training code at \url{github.com/ETH-PBL/elmap-rl-controller}.

\end{abstract}




\section{Introduction} \label{sec:intro}

Small-scale quadrupedal robots are becoming increasingly viable for real-world applications \cite{quadrupedal_review}. Their smaller size and lighter weight make them easier to integrate into human environments and less prone to cause harm during interactions \cite{dudzik2020robust}. Nevertheless, their limited computational resources present challenges in achieving robust, real-time locomotion on uneven terrains \cite{extero_egocentric}.

Reinforcement learning (RL)-based controllers have emerged as a promising solution for robust quadrupedal locomotion \cite{zhang2022deepreinforcementlearningforreal}.
Exteroceptive controllers, in contrast to those relying solely on proprioception, adapt better to complex terrains by providing richer environmental context through external sensors like depth cameras and LiDARs. However, they face challenges due to sensor noise and increased computational overhead \cite{proprio_rsl, extero_rsl}. 
Elevation mapping has become a popular strategy for tackling these limitations, allowing for terrain mapping from sensor data \cite{elevation_mapping_gpu, rsl_elev_uncertainty}. Nevertheless, elevation maps can remain sensitive to odometry errors and sensor noise, which can introduce drift and degrade performance \cite{extero_rsl,extero_egocentric}. Although these issues may be tackled by training with simulated noise, this tends to increase the RL task difficulty.
To address this, some approaches implement complex two-phase training architectures \cite{extero_rsl}.

 \begin{figure}[t]
    \centering
    \includegraphics[width=\columnwidth]{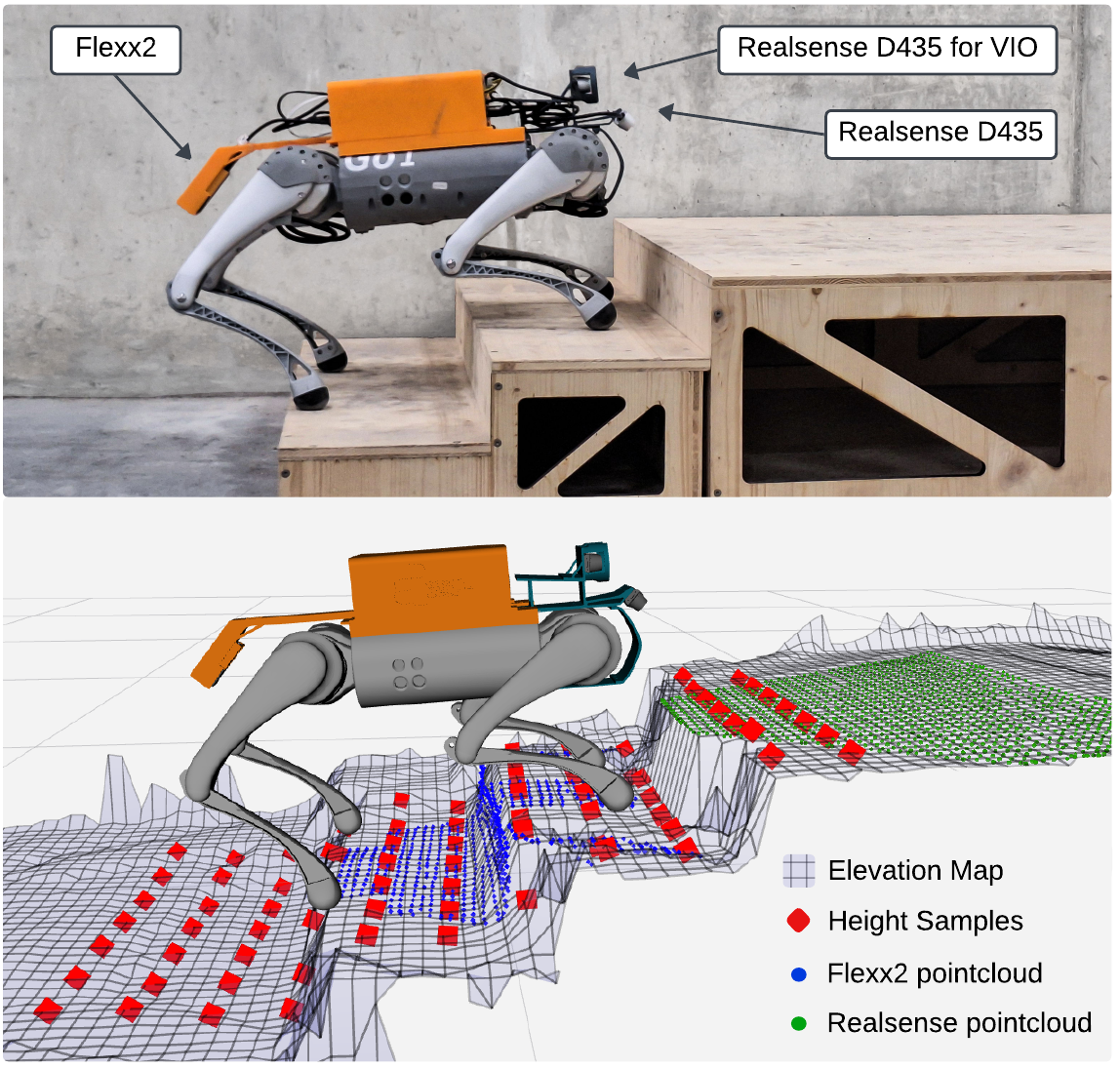}
    \caption{
    Top: Real-world stair climbing experiment with annotated sensors. Bottom: Visualization of the robot's state, showing filtered point clouds from the depth sensors, the elevation map, and the height samples passed to the policy.
    }
    \label{fig:title_page}
\end{figure}

Instead, we propose a robust real-time exteroceptive controller relying on elevation mapping combining stereo and time-of-flight (ToF) camera data, whilst exploiting a concurrent training approach proposed by \cite{concurrent_training} for proprioceptive controllers, resulting in a simplified single-phase RL architecture. We leverage the simultaneous training of both the control policy as well as an estimator, which acts as an odometry source for elevation mapping, optionally fused with visual-inertial odometry (VIO). This allows the robot to choose between computationally expensive, highly accurate odometry, or a lightweight alternative depending on resource constraints. Specifically, we summarize the following contributions:
\begin{itemize}
    \item A robust exteroceptive RL-based controller tailored to a resource-constrained small quadrupedal robot that balances performance and computational efficiency, leveraging concurrent training of both policy and state estimator.

    \item A minimal sensor configuration for elevation mapping, composed of both a stereo and a ToF camera, which enables robust performance without VIO. 
   
    \item A comprehensive evaluation and benchmarking, including ablation studies, to validate the system's real-world performance.
\end{itemize}



\section{Related Work} \label{sec:related}

Recent advances in reinforcement learning (RL)-based locomotion controllers have demonstrated robust performance in quadrupedal locomotion, including controllers that directly output setpoints for low-level proportional-derivative (PD) joint controllers \cite{proprio_rma,walktheseways,extero_egocentric,rl_rapid_loco} and can operate reliably even at low control frequencies \cite{rl_low_freq}. These can be divided into proprioceptive controllers, which use internal sensors like inertial measurement units (IMUs) and joint encoders \cite{proprio_rsl,proprio_rma,concurrent_training, walktheseways}, and exteroceptive controllers that integrate external sensors such as depth cameras and LiDAR \cite{extero_rsl,extero_egocentric}.
Exteroceptive systems offer better adaptability to complex terrains but suffer from sensor noise and computational overhead \cite{extero_egocentric}.
Recent parkour-based exteroceptive locomotion approaches \cite{parkour_corl, parkour_extreme, parkour_rsl} adopt highly dynamic motions but as a result, lack accurate velocity tracking.
While end-to-end navigation approaches fusing planning with locomotion have shown potential \cite{navigation_1, navigation_2}, many systems still benefit from velocity-tracking low-level locomotion policies \cite{navigation_3, navigation_4}.

To address sim-to-real challenges, extensive domain randomization and sensor noise simulation are employed, rendering RL training more difficult. Most approaches adopt two-phase training: an initial phase with privileged noiseless data, followed by training with simulated sensor noise \cite{extero_rsl, extero_confined_rsl,proprio_rma,extero_egocentric}. Another promising approach proposed in \cite{concurrent_training} is concurrent training of the policy and a state estimator, which simplifies the training pipeline.

Depth sensors such as depth cameras and LiDARs, commonly used for exteroceptive perception, are affected by noise and generate a large quantity of data.
Due to their computational efficiency, single-camera end-to-end approaches have been mostly employed for small-scale robots, however, remain subject to sim-to-real challenges \cite{extero_egocentric,parkour_corl,parkour_extreme}.
Other approaches consider elevation mapping \cite{rsl_isaac_gym,extero_rsl,extero_confined_rsl}, which fuses data from multiple sensors to provide a more consistent environmental model, that can also be extended to autonomous navigation systems \cite{rsl_elev_prob, rsl_elev_uncertainty}. Elevation maps are however resource-intensive and susceptible to odometry drift, though a recent GPU-accelerated implementation demonstrates real-time viability on Nvidia Jetson platforms \cite{elevation_mapping_gpu}. To compensate for the elevation mapping drift, two-phase training has been employed by adding realistic noise, resulting in a robust policy \cite{extero_rsl}.

Elevation mapping requires accurate odometry, which is challenging, particularly for RL-based controllers on rough terrain.
Classical and learning-based proprioceptive estimators have been proposed \cite{odom_proprio,odom_learned,concurrent_training}, while more robust odometry systems require fusing VIO or LiDAR resulting in significant computational overhead \cite{cerberus, vilens}.

Hence, there exists an opportunity to leverage concurrent training of the exteroceptive policy and state estimator, using it as an odometry source for efficient elevation mapping and robust locomotion in resource-constrained robots.





\section{Methodology} \label{sec:method}

Our system is built around a modified Unitree Go1 quadrupedal robot, equipped with additional hardware and sensors, as shown in \autoref{fig:title_page}. The robot is outfitted with an Intel NUC with a 10th-generation i7 CPU NUC10i7FNKN and a Nvidia Jetson Orin Nano to handle both the computational needs of real-time navigation and locomotion control. For exteroception, we integrate a front-mounted Intel Realsense D435 stereo depth camera and a rear-mounted PMD Flexx2 ToF camera for elevation mapping. An additional Realsense D435 camera, used as an infrared (IR) camera stereo setup, is integrated for visual-inertial odometry (VIO). The Flexx2 camera complements depth perception, improving robustness in varying lighting conditions and reducing the risk of sensor failure due to reflective or low-texture surfaces. It is mounted at the rear and specifically covers the area beneath the robot as illustrated in \autoref{fig:title_page}, where drift in the elevation map poses a particular challenge for achieving robust locomotion.
This LiDAR-free sensor configuration aims to balance power, weight, size, and computational efficiency. 

\begin{figure}[t]
    \centering
    \includegraphics[width=\columnwidth]{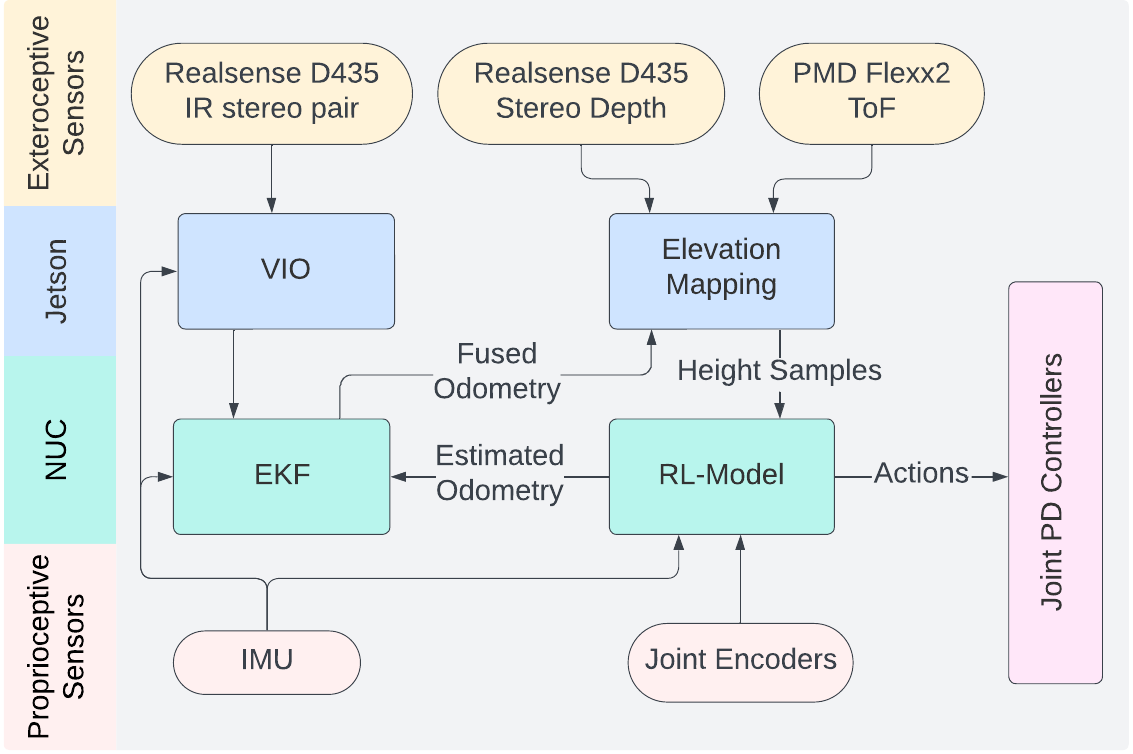}
    \caption{Deployed System overview: estimated odometry from the RL model is fused with IMU data and optionally VIO through an Extended Kalman Filter (EKF). The odometry along with the exteroceptive sensors are used to create an elevation map, out of which height samples are extracted and fed along with the proprioceptive sensors into the RL model, which outputs both actions and estimations. The Jetson Orin Nano processes the VIO and elevation mapping, while the Intel NUC handles model inference and EKF computation.}
    \label{fig:system_overview_flowchart}
\end{figure}

A system overview of all the components of the deployed controller is shown in \autoref{fig:system_overview_flowchart}.

\subsection{RL training} \label{subsec:method_rl_training}

\begin{figure*}[ht]
    \centering
    \includegraphics[width=0.95\textwidth]{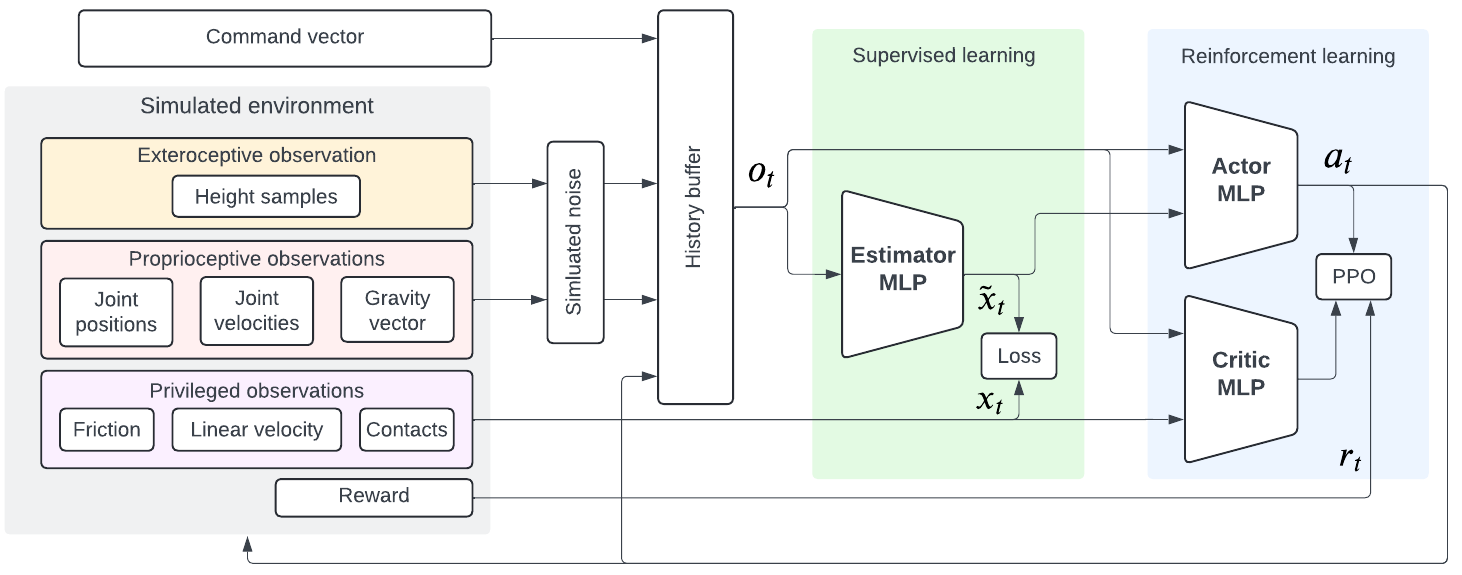}
    \caption{Concurrent training framework: the policy is optimized with PPO, and the estimator is trained via supervised learning. The observation vector $\mathbf{o_t}$ is constructed from a history buffer. The actor receives a concatenation of observations and estimated variables $\mathbf{\tilde{x}_t}$ and outputs actions $\mathbf{a_t}$, while the critic receives the observations concatenated with privileged variables $\mathbf{x_t}$, which are also used to train the estimator. Rewards $\mathbf{r_t}$ are provided by the simulated environment.}
    
    \label{fig:rl_architecture}
\end{figure*}

Our RL-based controller is designed to achieve robust locomotion and accurate command velocity tracking, while also providing an odometry source with an estimator module.
The policy and the state estimator are trained concurrently as in \cite{concurrent_training} in a single stage, instead of the commonly used two-stage teacher-student training procedure.

The controller architecture is depicted in \autoref{fig:rl_architecture} and includes three Multi-Layer Perceptron (MLP) networks: actor, critic, and estimator. The actor and critic have a [512×256×64] structure, while the estimator has a [256×128] structure.
The actor and critic are trained using proximal policy optimization (PPO) while the estimator is trained via supervised learning using privileged simulated data. The actor's input is a concatenation of the observation vector and estimation variables, which consist of linear velocity, friction, and foot contact states, while privileged observations are used for the critic instead.
The command vector consists of forward and lateral linear velocities $v_x^{cmd},v_y^{cmd}$ as well as yaw velocity $\omega_z^{cmd}$. 
The observations include the command vector, proprioceptive data (joint positions, velocities, gravity vector obtained from IMU data), and exteroceptive data (height samples).
Height observations are sampled from the elevation map, at 77 evenly distributed positions in the  \qty{0.5}{m} x \qty{0.3}{m} region around the robot, as shown in \autoref{fig:title_page}.
To ensure robustness during real-world deployment, missing regions in the elevation map are filled with default height values based on the robot's reference frame.
Because of the introduced observation noise and dynamics randomization, the state becomes only partially observable \cite{extero_rsl}. To mitigate this, the 10-step history of all observed variables is concatenated into the observation vector, corresponding to a memory of \qty{200}{\ms}.
The policy actions are the setpoints of the low-level joint PD controllers, with low gains of $K_p=20$ and $K_d=0.5$, which have been shown to have a similar effect of torque control \cite{dynamics_rand}.

To bridge the sim-to-real gap, we apply extensive domain randomization, including friction, restitution, robot mass, and motor strengths, and inject observation noise.
To handle the elevation mapping failure modes, we apply realistic noise modalities from \cite{extero_rsl} to the height samples: per-sample Gaussian noise at each step, and a constant Gaussian $x, y, z$ bias resampled every \qty{7}{s}.
The policies are trained in an environment with discrete steps of randomized size up to \qty{30}{cm} high. While this approach can result in longer training times compared to using a terrain curriculum as in \cite{rsl_isaac_gym}, it avoids the need for curriculum design, simplifying the tuning of other parameters.
We use 20-second episodes with a randomized initial state, sampling command variables at the start and midway through each episode. Episodes are terminated early in cases of collisions between the terrain and the robot's trunk.

Reward terms, shown in \autoref{tab:reward_terms} are mostly adapted from \cite{rsl_isaac_gym}, with the addition of terms penalizing foot slip as well as both joint positions and trunk height deviations from a default configuration, to achieve a consistent walking gait. Additionally, high torques are heavily penalized to prevent the robot from executing rapid, high-torque movements that could lead to damage.

\begin{table}[ht]
\centering
\caption{Reward Terms and Weights}
\label{tab:reward_terms}
\begin{tabular}{lcl}
\toprule
\textbf{Reward Term}          & \textbf{Definition}                                    & \textbf{Weight} \\
\midrule
Lin. vel. track               & $\phi(\mathbf{v}_{xy}^{cmd} - \mathbf{v}_{xy})$          & $1.0$ \\
Ang. vel. track               & $\phi(\omega_{z}^{cmd} - \omega_{z})$                    & $0.5$ \\
Feet air time                 & $\sum_{f=0}^{4}(\mathbf{t}_{air}  - 0.25)$               & $3.0$ \\
Lin. $v_z$ penalty            & $-v_{z}^2$                                    & $-2.0$ \\
Ang. vel. penalty        & $-\|\mathbf{\omega_{xy}}\|^2$                                   & $-0.05$ \\
Joint position                  & $\sum\left(\mathbf{q} - \mathbf{q}_{\text{default}}\right)^2$ & $-0.1$ \\
Joint acceleration            & $-\|\mathbf{\ddot{q}}\|^2$                         & $-2.5 \times 10^{-7}$ \\
Joint torques                 & $-\|\mathbf{\tau}\|^2$                                      & $-0.0002$ \\
Action rate                   & $-\|\mathbf{a}_t - \mathbf{a}_{t-1}\|^2$               & $-0.01$ \\
Joint collisions                    & $-n_{coll}$                                     & $-1.0$ \\

Trunk height                   & $(h_{\text{trunk}} - h_{\text{default}})^2$              & $-5.0$ \\
Joint torque limits            & $\sum \max(0, |\tau| - \tau_{\text{limit}})$ & $-10.0$ \\
\bottomrule
\end{tabular}
\vspace{0.1cm}
\begin{flushleft}
\small
\hspace{0.7cm}
 $\mathbf{v},\mathbf{\omega}$: linear and angular velocities, $\mathbf{\tau}$: torques, \\ 
\hspace{0.7cm} $\mathbf{q}$: joint positions, $\mathbf{t}_{\text{air}}$: foot air time, \\ 
\hspace{0.7cm} $n_{\text{coll}}$: number of collisions,
$\phi(x) = \exp\left(-\frac{\|x\|^2}{0.25^2}\right)$.
\end{flushleft}
\end{table}

At each timestep, the total reward is calculated as the weighted sum of all reward terms. If the sum is negative, it is scaled by a factor of 0.25 to prevent penalties from dominating, ensuring that the policy remains focused on the velocity-tracking task.

The models were trained for 60k iterations in Isaac Gym with 4096 parallel environments on an Nvidia RTX 4090 GPU, resulting in 15 hours of training time.

The policy and estimator run at \qty{50}{Hz}, which is sufficient for robust locomotion when using PD controllers, as demonstrated in \cite{rl_low_freq}.
Both the estimator and the policy networks are deployed on the Intel NUC, and achieve an average inference time of \qty{1.83}{\ms} including elevation map sampling. This remains well below the \qty{20}{\ms} inference window, allowing the CPU to remain largely available for other applications.


\subsection{Elevation mapping}
\label{subsec:method_elevation}

The elevation mapping system uses the open-source GPU accelerated implementation from \cite{elevation_mapping_gpu}, combining point cloud data from the downward-facing Realsense D435 and PMD Flexx2 cameras to generate a real-time terrain model.

The two sensors generate point clouds at \qty{30}{Hz}, which are then filtered.
Point cloud preprocessing includes removing outliers and applying a robot body filter that masks out the robot's legs using its  model and joint encoder data, preventing mapping errors caused by the robot's own body. To reduce computational load, the point clouds are also downsampled to a maximum spatial resolution of \qty{0.025}{m}.

The elevation map uses a grid-based 2.5D representation with a resolution of \qty{0.025}{\m} and a maximal size of \qty{5} x \qty{5}{m}. As soon as a grid cell receives new measurements, its height value is updated by using a Kalman filter which considers variance in terms of the sensor and time.
To address odometry drift, the framework compensates by calculating the average discrepancy between current sensor measurements and the elevation map, applying a global z-shift based on the error. 
While this can mitigate minor drift, it is insufficient for significant drift and the elevation map quality is still vulnerable to sensor noise.

The elevation mapping framework is executed in real-time at \qty{30}{\Hz} on the Jetson Orin Nano.

\subsection{Odometry estimation}

A robust odometry fusion framework integrates linear velocity estimates from the RL-trained estimator with IMU orientation data in an extended Kalman filter (EKF), optionally also incorporating a VIO source.
This approach ensures robustness against VIO failure by providing reliable fallback odometry through the estimator model.
Although less precise due to the sim-to-real gap, the estimator maintains functionality without catastrophic failures and the system can operate entirely without VIO, freeing up computational resources.
Compared to tightly coupled fusion methods  \cite{vilens,cerberus}, this setup is more modular and resource-efficient.

For VIO, we use the GPU-accelerated Nvidia Isaac ROS Visual SLAM \cite{isaac_ros} running on the Jetson Orin Nano along with elevation mapping, running at 90 FPS using the infrared stereo camera setup of a Realsense D435 at 640x360 resolution.



\section{Experiments \& Results} \label{sec:results}

The proposed controller is accurately evaluated in different configurations to demonstrate its performance and an ablation study is presented to assess the contribution of individual components.
Each of the benchmarked controller configurations consists of three main components. First, a model consisting of a trained policy and estimator, with the \textit{proposed model} being our best. Second, the odometry configuration: whether to use the full EKF setup, fusing VIO, estimated odometry and IMU, or just estimated odometry and IMU. Third, the sensor configuration: with or without the rear-mounted Flexx2. When not specified, the configuration includes the Flexx2 and uses VIO in the odometry.

\subsection{RL training results} \label{subssec:results_training}

\begin{figure}[t]
    \centering
    \includegraphics[width=\columnwidth]{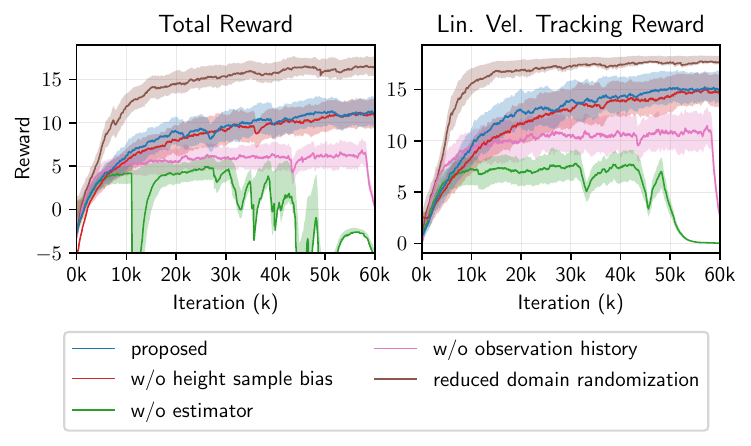}
    \caption{Comparison of total reward and linear velocity tracking reward across various models.}
    \label{fig:training_runs}
\end{figure}

The training performance of the proposed model is evaluated, comparing it against the following ablated versions:
\begin{itemize}
    \item \textit{w/o height samples bias}: the height sample bias noise is removed, leaving only the per-sample Gaussian noise.
    \item \textit{w/o estimator}: The estimator is removed.
    \item \textit{w/o observation history}: The observation history for both the estimator and policy is disabled.
    \item \textit{reduced domain randomization}: the extent and the magnitude of the domain randomization is reduced to match the one used in \cite{rsl_isaac_gym}.
\end{itemize}
All the models were trained for 60k iterations. \autoref{fig:training_runs} shows the learning curves for total reward and linear velocity tracking reward.

The proposed model shows consistent learning, with both the total and linear velocity rewards starting to plateau after 20k iterations, indicating a stable policy. After 60k iterations, no improvement in the real-world performance is observed.
The model without height sample bias performs similarly but with slightly lower rewards.
A possible explanation could be that the bias forces the model to adopt a more robust walking strategy early on.
Models trained without the estimator or observation history failed to perform consistently, confirming the importance of these components in handling the large domain randomization needed to reduce the sim-to-real gap.
Finally, the model trained with reduced domain randomization learns faster and achieves a higher final reward as expected, since the RL task becomes easier.

\subsection{Real-world step traversal}

\begin{figure}[t]
    \centering
    \includegraphics[width=\columnwidth]{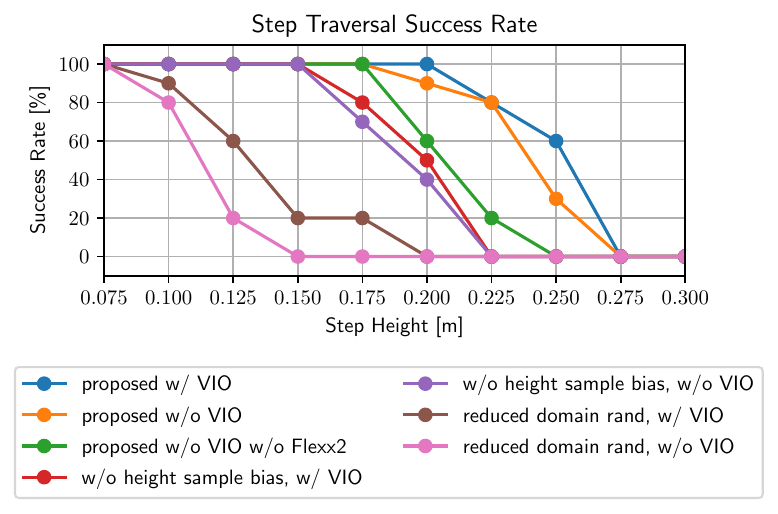}
    \caption{Step traversal real-world experiment, where the success rate of various configurations is plotted against step height.}
    \label{fig:step_traversal}
\end{figure}

We evaluate the controller robustness in real-world experiments which test the robot's ability to traverse steps of varying heights, from \qty{7.5}{\cm} to \qty{27.5}{\cm}, with success rates shown in \autoref{fig:step_traversal}. In each experiment, the robot is commanded with a forward velocity of \qty{0.5}{m/s} to climb and descend a single \qty{30}{\cm} deep wooden step. The experiment is deemed successful if the robot can walk from \qty{1}{\m} before the obstacle to \qty{1}{\m} after its end without falling, getting stuck, or being automatically stopped due to unsafe joint positions.

The full EKF configuration with VIO and the Flexx2 sensor performed best, being able to traverse steps up to \qty{25}{cm}. This corresponds to the physical limit due to the robot size while maintaining a walking gait \cite{extero_egocentric}. To achieve higher step heights a more dynamic motion is required \cite{parkour_corl,parkour_extreme}, which is not in the focus of this work.
Although without VIO the success rates for steps above \qty{20}{cm} decrease, it is still able to overcome obstacles of \qty{22.5}{cm} with 80\% success rate.
However, if the Flexx2 sensor is removed, the performance noticeably decreases. This shows that the addition of the Flexx2 sensor significantly mitigates the increase in odometry drift, justifying its inclusion.
These results demonstrate that the controller remains robust even in the event of VIO failure or deactivation, enhancing the versatility of the system.

The model trained without height sample bias is less robust and performs worse in the step traversal experiment than our proposed model, even when paired with better VIO odometry. Though the models perform similarly in training, real-world performance suffers, highlighting a larger sim-to-real gap. 
Moreover, the model trained with reduced domain randomization performs poorly in real-world tests, despite showing the best results in simulation, further underscoring the importance of addressing the sim-to-real gap.

\subsection{Elevation mapping quality} \label{subsec:results_elev_map_qual}

We evaluated the elevation mapping quality across different configurations. In the experiments, the robot was commanded to traverse a wooden obstacle at constant forward velocities of \qty{0.5}{m/s} in one trial and \qty{1.0}{m/s} in the other, with $v_{y}^{cmd}$ and $\omega_{z}^{cmd}$ used to maintain its trajectory. The obstacle comprised two \qty{10}{cm} high steps leading to a \qty{30}{cm} high platform, followed by a downward ramp.

The mapping quality was assessed by computing the error between the predicted elevation map and a ground-truth scan, focusing on the \qty{0.5}{m} x \qty{0.3}{m} region around the robot. The ground-truth terrain was generated using a Z+F 3D scanner, producing a grid map with a \qty{0.0175}{m} resolution. The robot's position and orientation were aligned with the terrain using a Vicon motion capture system.

The error was calculated using the one-way L1 Chamfer Distance, which measures the average Euclidean distance from each point in one cloud to the nearest point in the other:

\begin{equation} d_{cd}(S_1, S_2) = \sum_{x \in S_1} \min_{y \in S_2} |x - y|^2_2 \label{eq
} \end{equation}

The Chamfer Distance was computed at \qty{20}{\Hz} for each elevation map and averaged across all trials.

\autoref{Tab:elmap} presents the results from three different odometry sources. Ground-truth odometry (GT), provided by a Vicon motion capture system, was used as a baseline for comparison to ensure no influence from odometry inaccuracies. As expected, GT shows the lowest Chamfer Distance error and is minimally affected by the additional measurements from the Flexx2 sensor.
For the odometry setup using EKF with VIO, there is a noticeable decline in mapping quality due to odometry imperfections.
The Flexx2 sensor however, reduces the error by {13.24\%} compared to the configuration without the additional rear camera. When utilizing poorer odometry (without VIO), the Flexx2 sensor reduces the Chamfer Distance error by {28.56\%}.
Thus, the data clearly shows that the less accurate the odometry, the more beneficial the additional camera becomes.
Additionally, with Flexx2, there is only a {23.45\%} increase in error when not using VIO vs when using it, in contrast to {49.92\%}, indicating that the presence of the Flexx2 significantly reduces the drawbacks of having more noisy odometry.

\begin{table}[ht]
\centering
\caption{Mean Chamfer Distance Metric}
\label{Tab:elmap}
\begin{tabular}{cccc}
\toprule

\textbf{Odom Type} & \textbf{Flexx2} &
\textbf{Mean Chamfer Dist [cm] ↓} \\
\midrule
\multirow{2}{*}{GT}          & Yes & \textbf{1.155} \\
                             & No  & 1.166 \\    
\midrule
\multirow{2}{*}{EKF w/ VIO}  & Yes & 1.157 \\
                             & No  & 1.322 \\
\midrule
\multirow{2}{*}{EKF w/o VIO} & Yes  & 1.416 \\
                             & No   & 1.982 \\
\bottomrule

\end{tabular}
\end{table}

\subsection{Odometry accuracy} \label{subsec:results_odom}

We validate the odometry in the different configurations against ground truth using the same experiments conducted for elevation mapping evaluation.
The relative trajectory error (RTE) is computed following \cite{rpj_odom_eval} over \qty{1}{\m} trajectory segments to assess local accuracy, which directly affects the sampled region of the elevation map. The mean translation errors are presented in \autoref{Tab:odomtrack}.

VIO-only demonstrates the lowest error, while a slight degradation is observed when fusing it with the estimated odometry due to the latter's higher drift. However, the added robustness provided by the EKF against VIO tracking losses justifies its use, outweighing the minor reduction in accuracy.
When VIO is removed, the error increases by 43.11\%, which, though higher, demonstrates that the estimator still achieves effective zero-shot generalization.

\begin{table}[ht]
\centering
\caption{Relative Trajectory Errors}
\label{Tab:odomtrack}
\begin{tabular}{lcccc}
\toprule

\textbf{Odom Type} & { mean translation RTE [m]↓} \\
\midrule
EKF w/ VIO  & 0.0529  \\ 
EKF w/o VIO  & 0.0757  \\
VIO only  & \textbf{0.0480}  \\
\bottomrule

\end{tabular}
\end{table}

\subsection{Command velocity tracking}

As we aim to integrate the proposed model within an autonomous navigation pipeline, accurate command velocity tracking is crucial. We evaluate the following controller configurations: first, a baseline blind model trained solely on flat terrain up to 20k iterations specifically targeted towards the command velocity tracking task. Second, the proposed model with default height sample values as described in \autoref{subsec:method_rl_training}, mirroring a failure of the elevation mapping, and third, the proposed model and both with and without VIO.
We conduct a set of experiments collecting GT from the motion capture system and applying all velocity combinations in the following ranges:

\begin{equation}
\begin{aligned}
    v_{x}^{\text{cmd}} &\in \{-1.0, -0.5, 0, 0.5, 1.0\} \\
    v_{y}^{\text{cmd}} &\in \{-0.5, 0, 0.5\} \\
    \omega_{z}^{\text{cmd}} &\in \{-1.0, 0, 1.0\}
\end{aligned}
\end{equation}

Each combination is executed for 2 seconds, and the steady-state tracking performance is evaluated by calculating the root mean square (RMS) error of the measured data after a settling time of \qty{0.7}{\s}.
The overall root mean square error computed for every experiment in each velocity direction is summarized in \autoref{Tab:veltrack}.

The velocity tracking performance results intuitively show the baseline blind policy, trained specifically for command tracking, achieves the best performance. Nevertheless, tracking performance remains comparable across all three configurations of the proposed model, and in line with the performance of proprioceptive controllers from \cite{concurrent_training}.
The configuration using default height scan values, effectively mimicking a blind policy, shows impressive tracking performance. This demonstrates the policy's ability to maintain good tracking even without elevation map inputs. Interestingly, it performs slightly better than the configuration with VIO enabled, likely due to being unaffected by odometry drift.
This is particularly evident when the robot walks backward after accumulating elevation map drift, which worsens when relying solely on the estimated odometry.
Nonetheless, the proposed model demonstrates robustness, maintaining accurate tracking in all configurations.

\begin{table}[ht]
\centering
\caption{Command Velocity Tracking RMS Errors}
\label{Tab:veltrack}
\begin{tabular}{lccc}
\toprule
\textbf{Controller Configuration} & {$v_x$ [m/s]↓} & {$v_y$ [m/s]↓} & {$\omega_z$ [rad/s]↓} \\
\midrule
\textit{Blind baseline} & \textbf{0.0758} & \textbf{0.1109} & \textbf{0.2659} \\
\textit{Proposed, w/o elevation map} & 0.1192 & 0.1535 & 0.2829 \\ 
\textit{Proposed, w/ VIO} & 0.1198 & 0.1682 & 0.2947 \\
\textit{Proposed, w/o VIO} & 0.1177 & 0.1768 & 0.2949 \\
\bottomrule
\end{tabular}
\end{table}



\section{Conclusion} \label{sec:conclusion}

This work presented an RL-based controller for quadrupedal locomotion in challenging environments on a small-scale robot.
Robust exteroception is achieved through real-time GPU-accelerated elevation mapping using LiDAR-free sensing from two depth cameras: stereo (Realsense D435) and ToF (PMD Flexx2). 
We adopt a simplified RL architecture by concurrently training a policy and state estimator providing an odometry source to be optionally fused with GPU-accelerated VIO. 
We perform an ablation study in order to evaluate the performance of the system both with and without VIO as well as with and without the Flexx2 camera.
The proposed controller successfully traversed steps up to \qty{17.5}{\cm} without failure and achieved a success rate of over  80\% for steps of \qty{22.5}{\cm}, both when utilizing VIO and when relying solely on the estimated odometry.
This shows that incorporating the Flexx2 camera effectively preserves traversal performance, despite the 43.11\% increase in mean translation RTE when VIO is removed.
Additionally, the results indicate that the Flexx2 sensor improves elevation mapping accuracy, reducing the Chamfer Distance error by 28.56\% when VIO is unavailable, thereby mitigating the impact of odometry drift.
The proposed method not only introduces redundancy against VIO failures but also allows for the option to free computational resources by disabling it. Finally, we show that the proposed controller is able to accurately track forward and yaw velocities of up to \qty{1.0}{m/s} and \qty{1.5}{rad/s} respectively, needed for the successful integration in an autonomous navigation pipeline.

While the controller shows robustness against odometry drift and performs well on hard and visible terrains, its performance under depth sensor failure, particularly in cases with invisible or highly deformable objects, remains to be evaluated.

This work contributes to the advancements in small-scale autonomous robots by laying a solid foundation for reliable, resource-efficient quadrupedal locomotion.


\section*{Acknowledgment}

This work was supported by the ETH Future Computing Laboratory (EFCL). 
We would like to thank Hilti for assisting with the 3D
pointcloud acquisition.


\balance
\bibliographystyle{IEEEtran}
\bibliography{./bib/main}

\end{document}